\documentclass[10pt,twocolumn,letterpaper]{article}

\usepackage{3dv}
\usepackage{times}
\usepackage{epsfig}
\usepackage{graphicx}
\usepackage{amsmath}
\usepackage{amssymb}

\usepackage{multirow,multicol,makecell}
\usepackage{acronyms}
\usepackage{esvect}

\threedvfinalcopy 

\def\threedvPaperID{243} 

\ifthreedvfinal\fi
\begin{document}

\title{Learning Rotation-Invariant Representations of Point Clouds Using Aligned Edge Convolutional Neural Networks}


\author{Junming Zhang\\
University of Michigan\\
{\tt\small junming@umich.edu}
\and
Ming-Yuan Yu\\
University of Michigan\\
{\tt\small myyu@umich.edu}
\and 
Ram Vasudevan\\
University of Michigan\\
{\tt\small ramv@umich.edu}
\and 
Matthew Johnson-Roberson\\
University of Michigan\\
{\tt\small mattjr@umich.edu}
}

\maketitle

\begin{abstract}
Point cloud analysis is an area of increasing interest due to the development of 3D sensors that are able to rapidly measure the depth of scenes accurately. 
Unfortunately, applying deep learning techniques to perform point cloud analysis is non-trivial due to the inability of these methods to generalize to unseen rotations.
To address this limitation, one usually has to augment the training data, which can lead to extra computation and require larger model complexity. 
This paper proposes a new neural network called the Aligned Edge Convolutional Neural Network (AECNN) that learns a feature representation of point clouds relative to Local Reference Frames (LRFs) to ensure invariance to rotation. 
In particular, features are learned locally and aligned with respect to the LRF of an automatically computed reference point. 
The proposed approach is evaluated on point cloud classification and part segmentation tasks.
This paper illustrates that the proposed technique outperforms a variety of state of the art approaches (even those trained on augmented datasets) in terms of robustness to rotation without requiring any additional data augmentation.
\end{abstract}
\section{Introduction}
The development of low-cost 3D sensors has the potential to revolutionize the way robots perceive the world. 
For this revolution to be realized, algorithms to interpret and classify the large volumes of point clouds generated by these sensors must be developed.
To construct such algorithms, one could be inspired by the successes of deep learning approaches that robustly interpret 2D images in the presence of noise or lighting, rotation, and scaling variability.
These deep learning approaches achieve impressive performance by relying on representations that enforce lighting, rotation, and scaling invariance. 
Unfortunately, the lack of a representation that is able to enforce rotation invariance has hindered the application of deep learning techniques to analyze point clouds.

\begin{figure}[t]
    \centering
    \includegraphics[width=\linewidth]{{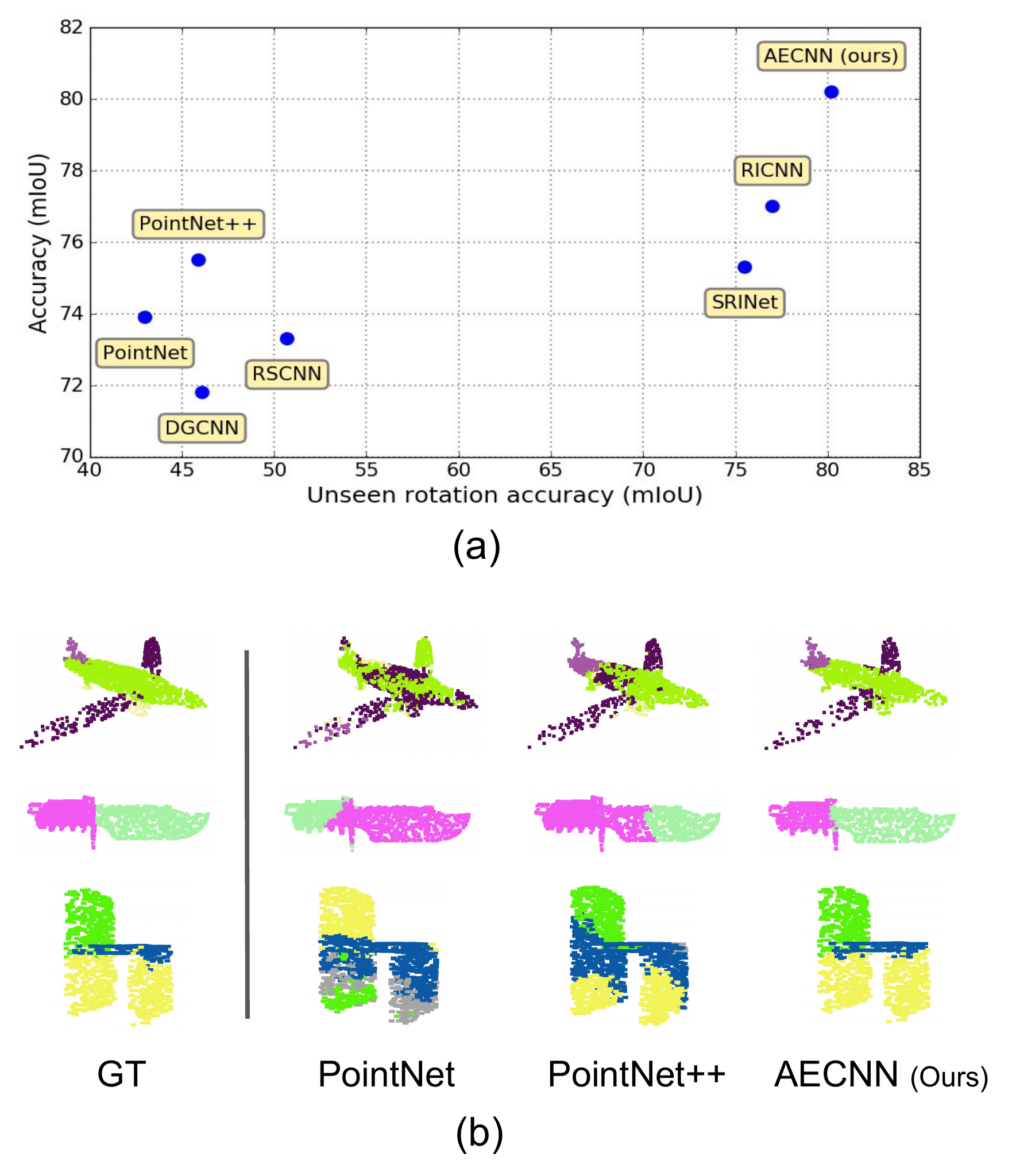}}
    \caption{An comparison of the performance of state-of-the-art techniques to the method developed in this paper while performing part segmentation on rotated point clouds from the ShapeNet dataset.
    We report results on models of arbitrary rotation during testing while they trained with only rotation along vertical direction ($x$ axis) and on models of arbitrary rotation during both training and testing ($y$ axis) (subfigure a).
    This degradation in performance can be seen on the quality of part segmentation in unseen views (subfigure b, where different colors correspond to different part categories).}
    \label{fig:seg_comparison}
\end{figure}

To address this challenge, researchers have typically converted point clouds into regular 2D~\cite{feng2018gvcnn,su2015multi,kanezaki2018rotationnet,su2018deeper} or 3D~\cite{voxnet,qi2016volumetric,wu20153d} grids before applying \acp{CNN} to learn a meaningful representation. 
Unfortunately, this conversion process degrades the resolution of measured objects which can adversely affect point cloud analysis.
More recently, PointNet~\cite{qi2017pointnet} was proposed to preserve some of this geometric information by using a symmetric kernel that could enforce permutation invariance.
This approach allowed a user to directly treat a point cloud as an input into a \ac{DNN} and get a global vector representing the input point cloud as an output. 
This work was subsequently extended in a variety of ways to preserve local structure within a point cloud that proved to be important while performing classification~\cite{qi2017pointnet++,wang2019dynamic,liu2019relation}.
However, the representations developed by these methods rely on an individual point's absolute position, which hinders their ability to develop algorithms that are invariant to rigid body transformations.
Fig.~\ref{fig:seg_comparison}, for instance, illustrates the deficiency of these methods when they are applied to perform part segmentation on views that are unseen during training.

Typically, one can address this limitation and improve the robustness of \acp{DNN} to rigid body transformations by augmenting the training set with additional examples.
However, this requires additional computation and increased model capacity. 
For instance, during classification, a model must learn a function that maps the same object under different rigid body transformations into a similar feature in a feature space.
Rather than augment the training set, other approaches have focused on developing representations that can preserve rotational symmetry by directly converting point clouds into a spherical voxel grid and then extracting rotation-equivariant features~\cite{esteves2018learning,rao2019spherical,you2018prin}. 
Unfortunately this conversion still sacrifices resolution that can adversely affect point cloud analysis. 
To remedy this loss of information, others have proposed to represent point clouds relative to a \ac{LRF}, which is determined by a local subset of a point cloud~\cite{chen2019clusternet,zhang2019rotation}.
Each point in a neighborhood of a constructed \ac{LRF} is represented with respect to that \ac{LRF} and a local feature is learned for each point set. 
Subsequently these local features are fused together to define global features. 
These \ac{LRF}-based learned representations are invariant to rotation; however, as we illustrate in this paper, the accuracy of techniques utilizing \ac{LRF}-based representation are only marginally better than those utilizing an absolute coordinate based representation for point cloud classification tasks. 
This is in part because the learned local features for a pair of points are not aligned before they are fused together.

To address the limitations of existing approaches, this paper proposes a novel 3D representation of point clouds that is invariant under rotation and introduces a new neural network architecture, the \ac{AECNN}, to utilize this representation.
As in prior work, we leverage the notion of \acp{LRF} to ensure that different orientations of a point cloud are mapped into the same representation. 
Each point in a neighborhood of a constructed \ac{LRF} is represented with respect to that \ac{LRF} before subsequent processing. 
This ensures that the model is able to learn internal geometric relationships between points rather than learning geometric relationships that are a function of the absolute coordinates of the points that may change after rotation.
Our proposed \ac{AECNN} architecture processes these local internal features and aligns them with local internal features drawn from other \acp{LRF} before fusing them together in a hierarchical fashion to define global features. 
Importantly, in contrast to prior work that utilizes a spherical coordinate system~\cite{chen2019clusternet} or non-orthogonal basis~\cite{sun2019srinet}, we construct a basis for the \acp{LRF} that is orthonormal.
This ensures that feature alignment can be computed in a straightforward manner which makes the hierarchical fusion of local features tenable.


The contributions of this paper are three-fold:
First, we propose a novel representation of points clouds that is invariant to arbitrary rotations.
Second, we propose a novel alignment strategy to align neighboring features within distinct \acp{LRF}. 
This makes it feasible to perform feature fusion within a hierarchical network, which makes a reasonable feature fusion between local and global features.
Finally, we illustrate that our propose representation is robust to rotation and achieves state-of-the-art results in both point cloud classification and segmentation tasks.

\section{Related work}
This section reviews the various techniques that have been applied to represent point clouds.

\textbf{View-Based and Volumetric Methods.} 
A variety of methods have represented 3D shape as a sequence of 2D images since they can leverage existing algorithms from 2D vision~\cite{he2016deep,he2017mask,long2015fully}.
These view-based methods typically project a 3D shape onto 2D planes from different views.
These different images are then processed by \acp{CNN}.
Though these methods achieve good performance at classification tasks using off-the-shelf architectures and pre-trained model \cite{feng2018gvcnn,su2015multi,kanezaki2018rotationnet,su2018deeper}, the projection of 3D shape onto 2D planes sacrifices 3D geometric structures that are critical during point cloud analysis. 

Converting point clouds into 3D voxels can preserve some of this geometric information. 
The higher the resolution of quantization, the more geometric information is preserved. 
These converted point clouds can then leverage existing \acp{CNN} with 3D kernels~\cite{voxnet,qi2016volumetric,wu20153d}. 
Since points are only sampled from the surface of objects, regular quantization can waste valuable resolution on the empty space within or outside objects.
Better partitioning methods, such as KD-tree~\cite{klokov2017escape} and Oct-tree~\cite{wang2017cnn,tatarchenko2017octree} have been proposed to address this limitation.
In contrast to these methods, our approach works directly with point clouds without requiring any conversion. 

\textbf{Point Set Learning.} 
Point set learning methods take raw point clouds as input.
The pioneering work in this area is PointNet~\cite{qi2017pointnet}, which independently transforms each point and outputs a global feature vector describing the input point cloud by aggregation using a max pooling layer. 
Unfortunately, PointNet is unable to learn local structure over increasing scales, which is important for high-level learning tasks.
Extensions that utilize hierarchical structure~\cite{qi2017pointnet++,li2018so}, graph network~\cite{wang2019dynamic,zhang2018graph}, or relation-aware features~\cite{liu2019relation} have been proposed to preserve local geometric structure during processing.
Other extensions that rethink the convolution operation to better accommodate point cloud processing have also  been proposed~\cite{wu2019pointconv,li2018pointcnn}. 
However, the representation that is learned by these approaches changes when the point clouds in the training set are rotated. 
As a result, these representations perform poorly when utilized during classification or segmentation tasks on rotated versions of point clouds that were not included during training.

\textbf{Rotation Learning.} 
Various methods have been proposed to either learn rotation-invariant or rotation-equivariant representations.
For instance, \acp{STN} have been proposed to learn rotation-invariant representations~\cite{qi2017pointnet}. 
The \ac{STN} learns a transformation matrix to align input point clouds without requiring that the alignment place the point clouds in some fixed ground-truth orientation.
As a result, the transformation matrix is not guaranteed to align objects to a consistent orientation which restricts its utility.
Other approaches achieve rotation invariance by relying on \acp{LRF} ~\cite{chen2019clusternet,zhang2019rotation,sun2019srinet}. 
However, the difficulty of aligning the features learned with respect to \acp{LRF}, as described earlier, has limited the potential expressive capabilities of these techniques. 
Since designing a model that is invariant to rotation is difficult, a variety of  methods have attempted to achieve rotation-equivariance using spherical convolutions~\cite{esteves2018learning,rao2019spherical,you2018prin}. 
Spherical convolutions require the spherical representation of a point cloud.
Unfortunately projecting 3D point clouds into 2D sphere results in a loss of information.

\section{Learning Rotation-Invariant Representation of Point Clouds}
This section introduces our proposed \ac{RIR} of point clouds using \acp{LRF}, our proposed aligned edge convolution designed for \ac{RIR}, and the proposed hierarchical network architecture for classification and segmentation tasks.

\begin{figure}[t]
    \centering 
    \includegraphics[width=\linewidth,height=0.4\linewidth]{{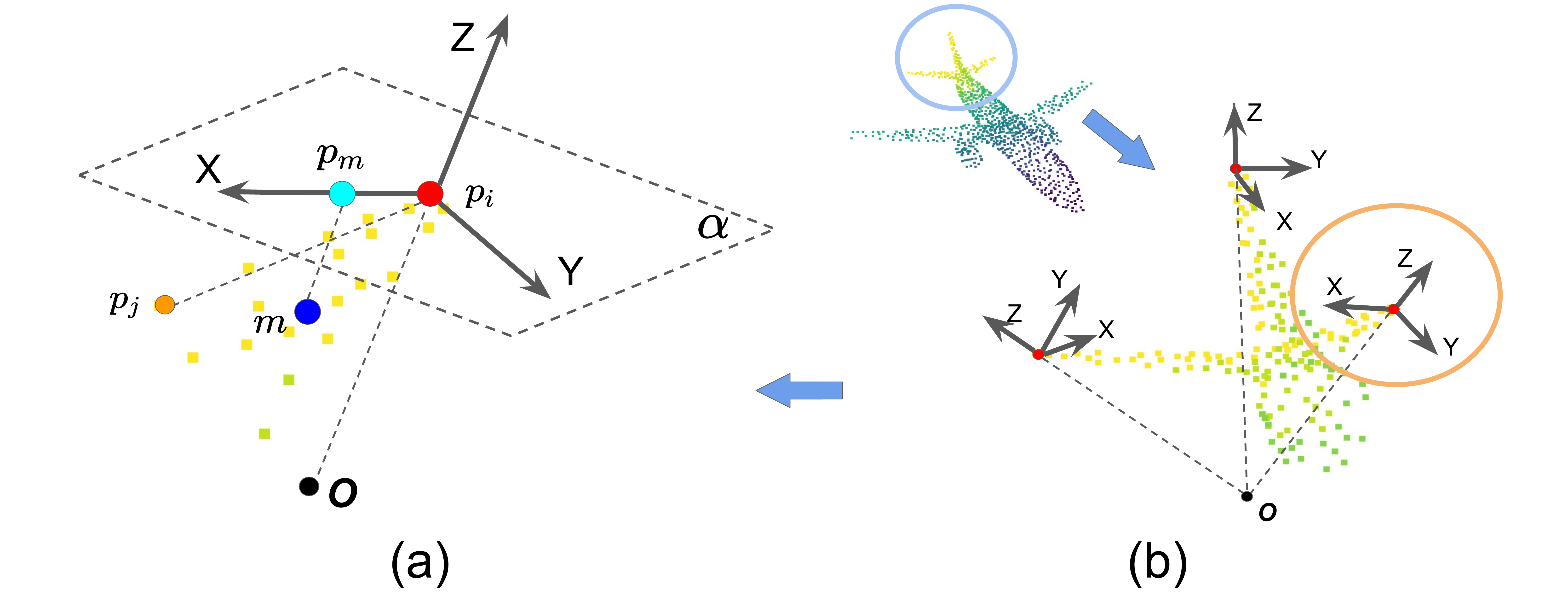}}
    \caption{
    An illustration of aligned edge convolution and the \acp{LRF} that help define it.
    To construct the \ac{RIR}, one takes the reference point $p_i$ and the $k$-nearest points to it (subfigure a). 
    The \ac{LRF} is determined by $p_i$ and the anchor point $m$ which is defined as the barycenter of the $k$-nearest points to $p_i$.
    The coordinates of $k$-nearest points are described with respect to the \ac{LRF}. 
    Note that \acp{LRF} may not be aligned due to independence of the local neighborhood of points that define each \ac{LRF} (subfigure b). 
    }
    \label{fig:LRF}
\end{figure}

\subsection{Rotation-Invariant Representation}
\label{section:RIR}

To construct a representation that is invariant to rotation, we represent a point's coordinates relative to a \ac{LRF}.
Each \ac{LRF} is defined using three orthonormal basis vectors and is designed to be dependent on local geometry, as is depicted in Fig. \ref{fig:LRF} (a). 
To define this \ac{LRF}, suppose we are given a reference point $p_i$ in a point cloud and a set of neighboring points $\{p_1,\ldots,p_k\}$ to $p_i$ in the point cloud whose coordinates are all described with respect to a coordinate system with global origin $o$.
Note, we describe how to select these reference points in Section \ref{subsec:network}, and neighboring points are those within a certain radius to the reference point.
Next, define an anchor point $m$ as the barycenter of the neighboring points:
\begin{equation}
    m = \frac{1}{k} \sum_{j=1}^{k} p_j,
\end{equation}
and the plane $\alpha$ that is orthogonal to $\vv{op_i}$ and intersects with $p_i$.
Using these definitions, we can define the projection of $m$ onto the plane $\alpha$:
\begin{equation}
    \vv{p_ip_m} = \vv{om} - \frac{\vv{op_i}}{|\vv{op_i}|} \cdot \langle \vv{om}, \frac{\vv{op_i}}{|\vv{op_i}|}\rangle.
\end{equation}
With this definition, we can construct the following coordinate axes for the \ac{LRF}:
\begin{equation}
\label{eq:1}
    \vv{x}:= \frac{\vv{p_ip_m}}{|\vv{p_ip_m}|}, \;\;\; \vv{z}:= \frac{\vv{op_i}}{|\vv{op_i}|}, \;\;\; \vv{y}:= \vv{z} \times \vv{x}.
\end{equation}
Note that $z$ axis is defined as the direction from global origin $o$ pointing at $p_i$; the $x$ axis is defined as the direction from $p_i$ pointing at $p_m$; and the $y$ axis is defined as the direction of cross product of $z$ and $x$ axis. 
The origin of \ac{LRF} is at the reference point $p_i$. We assume that the global origin $o$ is known, and in our case we use the center of point clouds.

We introduce rotation invariance by representing the set of neighboring points relative to their \ac{LRF}:
\begin{equation}
\label{eq:2}
    t^i_j = (\langle p_{ij}, \vv{x} \rangle, \langle p_{ij}, \vv{y} \rangle, \langle p_{ij}, \vv{z} \rangle)
\end{equation}
where $p_{ij} = p_j - p_i$ for each $p_j \in \{p_1,\ldots,p_k\}$.
Note $t^i_j$ is the \ac{RIR} for the point $p_j$ relative to the LRF at point $p_i$. 
We then use a PointNet structure to capture the geometry within the neighboring points using the \ac{RIR}:
\begin{equation}
    f(\{t^i_1, ..., t^i_k\}) = MAX(\{h(t^i_1), ..., h(t^i_k)\})
\end{equation}
where $f$ is a learning function, which takes a point set as input, and outputs a feature vector representing input point clouds, $h$ is a feature transformation function and is approximated by a \ac{MLP}. 
Note, the max pooling layer aggregates information. 

\subsection{Aligned Edge Convolution}

To capture the geometric relationship between points in a point cloud, the notion of edge convolution via DGCNN has been developed~\cite{wang2019dynamic}. 
To understand how edge convolution works, suppose we are given a point cloud with $n$ points, denoted by $P = \{p_1, ..., p_n\} \subseteq {\rm I\!R}^3$, along with F-dimensional features corresponding to each point, denoted by $X = \{x_1, ..., x_n\} \subseteq {\rm I\!R}^F$. 
Suppose we construct the $k$-\ac{NN} graph $(V, E)$ in the feature space, where $V={1, ..., n}$ and $E\subseteq V \times V$ are vertices and edges, then the edge convolution output at $i$-th vertex is given by:
\begin{equation}
\label{eq:6}
    x'_i = \underset{j:(i,j)\in E}{MAX} \; g(x_i, x_j - x_i)
\end{equation}
where $g$ is a \ac{MLP}.

Note, edge convolution is essentially performing feature fusion. 
It fuses the global information captured by $x_i$ with local neighborhood information captured by $x_i - x_j$. 
To be able to do this,  $x_i$ and $x_j$ must be learned in the same coordinate system. 
Unfortunately, it is nonviable to directly apply edge convolutions to features in our case. 
This is because in our case, $x_i$ and $x_j$ are learned relative to two different \acp{LRF}, and the \acp{LRF} of $x_i$ and $x_j$ may not be aligned, as is shown in the Fig. \ref{fig:LRF} (b).
As a result, applying edge convolution directly on our learned features may create inconsistent features.

To resolve this problem, we propose aligning $x_j$ into the \ac{LRF} of $x_i$ before performing feature fusion.
We call our approach, which is depicted in Fig. \ref{fig:architecture}, Aligned Edge Convolution (Aligned EdgeConv). 
To construct our approach, we begin by understanding the relationship between different \acp{LRF} which can be described using a rotation $R$ and translation $T$.
To construct this rotation and translation, suppose the basis of the \ac{LRF} for each feature is denoted by $E = \{e_1, ..., e_n\} \subseteq {\rm I\!R}^{3\times3}$.
Then the rotation matrix and translation vector can be computed by:
\begin{equation}
    R_j = e_i \cdot e_j^{-1} = e_i \cdot e_j^\top
\end{equation}
\begin{equation}
    T_j = t^i_j
\end{equation}
where $e_j$ is an orthogonal matrix defined in \eqref{eq:1} and $t_j$ is defined in \eqref{eq:2}. 

$R$ and $T$ describe the relationship between \acp{LRF}, so we use them to transform $x_j$ into the \ac{LRF} of $x_i$. 
Though it is easy to invert a rotation and translation in 3D, extending it to the high dimensional feature space  that $x_j$ lives in would be challenging. 
One option to resolve this problem is to apply an approach similar to the \ac{STN} proposed in the PointNet wherein one predicts a transformation matrix from $R$ and $T$ and applies it to $x_j$:
\begin{equation}
\label{eq:3}
    \hat{x}_j = \phi(R_j, T_j) \cdot x_j
\end{equation}
where $\phi$ is a \ac{MLP} and outputs an $F \times F$ matrix compatible with $x_j$. 
Typically a regularization term is added to the loss during training to constrain the feature transformation matrix to be close to an orthogonal matrix. 
Another option is to take $R$, $T$ and $x_j$ as inputs and directly output a transformed feature:
\begin{equation}
\label{eq:4}
    \hat{x}_j = \phi(R_j, T_j, x_j)
\end{equation}

In this paper, we utilize the second option. 
As we show in Section \ref{section:aligned_edgeconv}, option one requires more \acp{GPU} memory and has more parameters. 
Therefore we update the \eqref{eq:6} by:
\begin{equation}
    x'_i = \underset{j:(i,j)\in E}{MAX} \; q(x_i, \hat{x}_j - x_i).
\end{equation}
Similar to PointNet++~\cite{qi2017pointnet++}, we also include the \ac{RIR} $t^i_j$ in the edge convolution to maintain more information.
So the aligned edge convolution is given by
\begin{equation}
    x'_i = \underset{j:(i,j)\in E}{MAX} \; q(x_i, \hat{x}_j - x_i, t^i_j).
\end{equation}

\subsection{Network Architecture}
\label{subsec:network}

\begin{figure*}[t!]
    \centering
    \includegraphics[width=1\linewidth]{{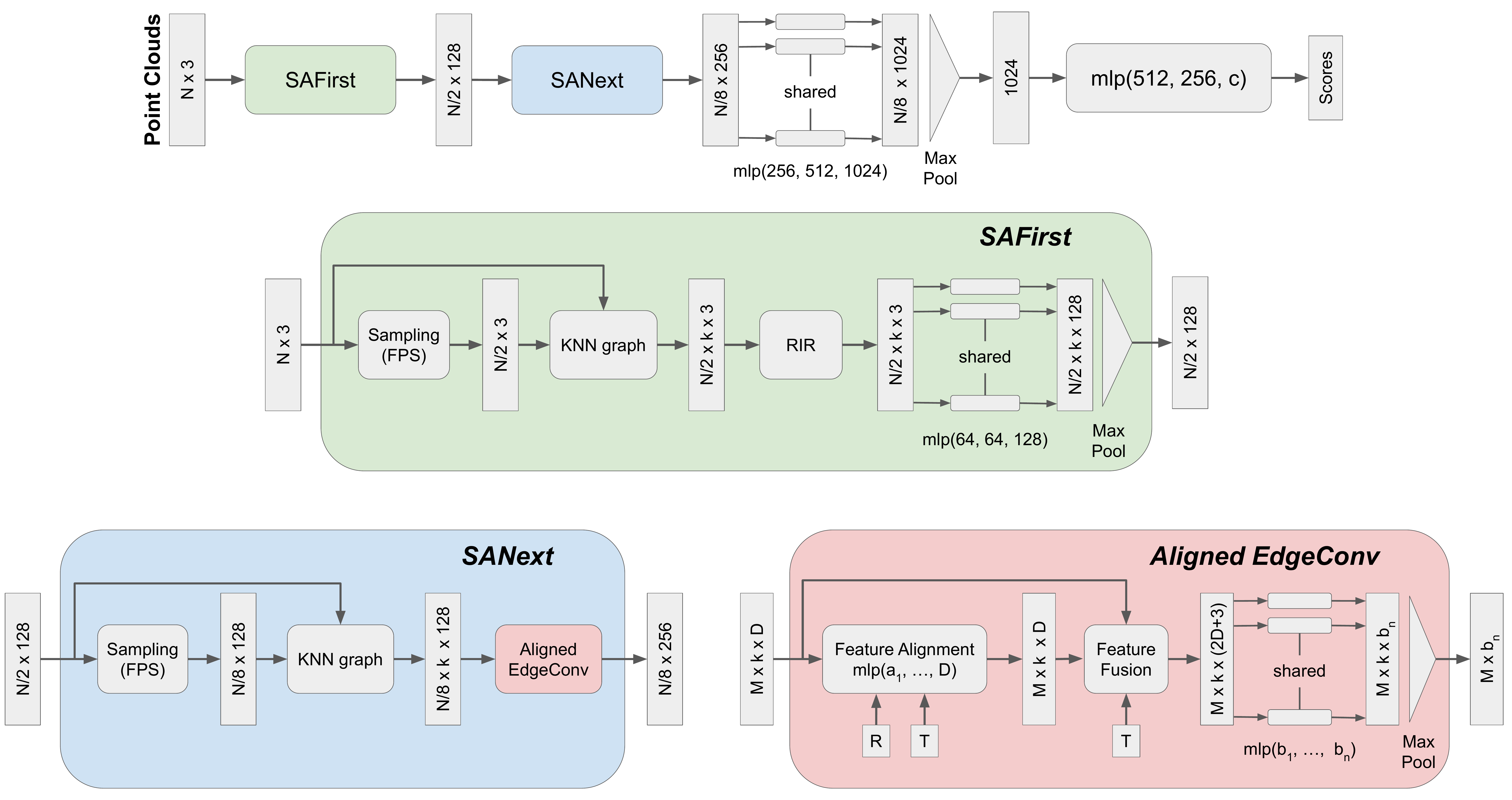}}
    \caption{An illustration of the deep hierarchical architecture proposed in this paper to learn a rotationally invariant representation of point clouds for  classification. 
    The network takes $N$ points as inputs and uses two \ac{SA} blocks to hierarchically learn a representation of larger and larger regions. 
    The final part aggregates features from the last \ac{SA} block and outputs a feature vector encoding the input point set. 
    The SAFirst block samples reference points and builds a $k$-\ac{NN} graph in Euclidean space. 
    It converts points into the \ac{RIR} and transforms them using a shared-weights PointNet structure that outputs feature vectors encoding information for each reference point' neighbors. 
    The SANext block applies a similar process to the SAFirst block except for building the $k$-\ac{NN} graph in feature space and extracting features using Aligned EdgeConv. 
    The feature alignment module within Aligned EdgeConv aligns local features to the \ac{LRF} of the reference point using the rotation matrix $R$ and the translation vector $T$ that are defined in the Section \ref{section:RIR}. 
    Then the aligned local features are fused with the feature at the reference point along with translation vector $T$. Note mlp in the figure is abbreviated for \ac{MLP}.
    The output is classification scores for $c$ classes.}
    \label{fig:architecture}
\end{figure*}

Our proposed network architecture that takes raw point clouds as input and learns a representation is depicted in Fig. \ref{fig:architecture}. 
Our architecture is inspired by techniques that perform local-to-global learning which has been successfully applied to 2D images~\cite{zhao2017pyramid} and has been shown to effectively extract contextual information. 
We exploit a hierarchical structure to learn both local and global representation. 
Our approach captures larger and larger local regions using two \ac{SA} blocks, proposed by PointNet++~\cite{qi2017pointnet++}, and the global features are constructed by aggregating outputs from the last \ac{SA} block using max pooling.

The two \ac{SA} blocks each have distinct structures; however, they share the same processing pipeline: sampling, grouping, and processing. 
The structure of the first \ac{SA} block (SAFirst) is illustrated in the green block in Fig. \ref{fig:architecture}.
Given the input point clouds, we use farthest point sampling (FPS) to subsample the point cloud while preserving its geometric structure.
The constructed points serve as the reference points during the construction of a $k$-\ac{NN} graph. 
The $k$-\ac{NN} graph is computed in Euclidean space, and it is used to generate the \ac{RIR} described in Section \ref{section:RIR}. 
A shared-weights PointNet structure then processes the \acp{RIR} for each local set of points and outputs a feature vector describing the set of points near each reference point.

The structure of SANext is shown within the blue block in  Fig. \ref{fig:architecture}. 
The sampling and grouping strategy in SANext is identical to the one in SAFirst except that one quarter the number of points are selected as reference points and the $k$-\ac{NN} graph is dynamically updated and computed in the feature space, which has been shown to be more beneficial and have larger receptive fields than a fixed graph version~\cite{wang2019dynamic}. 
Then the proposed aligned edge convolution extracts features encoding larger local regions than the previous \ac{SA} block. 
The rotation matrix $R$ and translation vector $T$, which can be derived from basis and positions of \acp{LRF}, are fed into a feature alignment module to align local features to the frame of reference point.
Essentially, the SANext is the building block that can be used to iteratively capture larger and larger local regions. 

Note for the segmentation task, we require a feature for each point. 
We adopt a similar strategy to PointNet++~\cite{qi2017pointnet++}, which propagates features from subsampled points to original points.
Specifically, the interpolated features from the previous layer are concatenated with skip linked features output from \ac{SA} blocks. 
The interpolation is done via the inverse distance weighted average based on $k$-\ac{NN}. 
Importantly, the proposed feature alignment idea is also integrated in this feature propagation pipeline.

\section{Experiments}
This section describes how we implement the network described in Section \ref{subsec:network} and how we validate its utility. 
First, we evaluate our method on a shape classification task (Sec \ref{section:modelnet40}) and part segmentation task (Sec \ref{section:shapenet}). 
Second, we evaluate the design of our \acp{LRF} (Sec \ref{section:LRF_validation}) and illustrate the effectiveness of the proposed aligned edge convolution (Sec \ref{section:aligned_edgeconv}).

\subsection{Implementation details}
\label{section:implementation}
We implement our network in PyTorch. 
All experiments are run on a single NVIDIA Titan-X GPU. 
During optimization, we use the Adam optimizer with batch size 32. 
Models are trained for 250 epochs. 
The learning rate starts with $1e{-3}$ and scale by $0.2$ every $100$ epochs. 

In some experiments, we augment the dataset with arbitrary rotations. 
However, it is impossible to cover all rotations in 3D space. 
Similar to ClusterNet~\cite{chen2019clusternet}, we uniformly sample possible rotations. 
Each rotation is characterized by a rotation axis $v$ and a rotation angle $\theta$ that is given by:
\begin{equation}
    R = I + (\sin{\theta})K + (1 - \cos{\theta})K^2
\end{equation}
where $K$ denotes the cross-product matrix for the rotation axis $v$ which has a unit length and $I$ is the identity matrix. 
In the experiments, we sample 3-dimensional vectors from a normal distribution and normalize $v$ to be a unit vector. 

We follow the approach presented in~\cite{esteves2018learning} to perform experiments in three different settings: 1) training and testing with rotation along the vertical direction (Y/Y), 2) training with rotation along vertical direction and testing with arbitrary rotation (Y/AR) and 3) performing arbitrary rotation during training and testing (AR/AR). 
The last two settings in particular are used to evaluate the generalization ability of the model under unseen rotations.

\begin{table*}[t]
\begin{center}
\begin{tabular}{lccccc}
\hline
Method & Inputs & Input size & Y/Y & Y/AR & AR/AR \\
\hline
MVCNN 80x~\cite{su2015multi} & view & $80 \times 224^2$ & 90.2 & 81.5 & 86.0 \\
Spherical CNN~\cite{esteves2018learning} & voxel & $2 \times 64^2$ & 88.9 & 76.9 & 86.9\\
\hline
PointNet~\cite{qi2017pointnet} & point & $1024 \times 3$ & 88.5 & 21.8 & 83.6 \\
PointNet++~\cite{qi2017pointnet++} & point & $1024 \times 3$ & 89.3 & 31.7 & 84.9 \\
RS-CNN~\cite{liu2019relation} & point & $1024 \times 3$ & 89.6 & 24.7 & 85.2\\
DG-CNN~\cite{wang2019dynamic} & point & $1024 \times 3$ & 91.7 & 31.5 & 88.0 \\
\hline
RI-CNN~\cite{zhang2019rotation} & point & $1024 \times 3$ & 86.5 & 86.4 & 86.4\\
SRINet~\cite{sun2019srinet} & point & $1024 \times 3$ & 87.0 & 87.0 & 87.0\\
ClusterNet~\cite{chen2019clusternet} & point & $1024 \times 3$ & 87.1 & 87.1 & 87.1\\
SF-CNN~\cite{rao2019spherical} & point & $1024 \times 3$ & \textbf{92.3} & 84.8 & 90.1\\
\hline
Ours & point & $1024 \times 3$ & 91.0 & \textbf{91.0} & \textbf{91.0} \\
\hline
\end{tabular}
\end{center}
\caption{Classification results on ModelNet40 dataset. We report the accuracy (\%) in three different settings: training and testing with rotation along the vertical direction (Y/Y), training with rotation along vertical direction and testing with arbitrary rotation (Y/AR), and performing arbitrary rotation during training and testing (AR/AR). Though our model is only the third best performer in the Y/Y setting it is the top model in each of the other categories. In particular our proposed model has superior performance in the Y/AR and AR/AR, which means that it can generalize well to unseen rotations.}
\label{table:classification}
\end{table*}

\subsection{Shape Classification}
\label{section:modelnet40}
One of the primary point clouds analysis tasks is to recognize the category of point clouds. 
This task requires a model to learn a global representation. 

\textbf{Dataset.} We evaluate our model on ModelNet40, which is a shape classification benchmark~\cite{wu20153d}. It provides 12,311 CAD models from 40 object categories, and there are 9,843 models for training and 2,468 models for testing. 
We use their corresponding point clouds provided by PointNet~\cite{qi2017pointnet}, which contain 1024 points in each point clouds. 
During training we augment the point clouds with random scaling in the range $[-0.66, 1.5]$ and random translation in the range $[-0.2, 0.2]$ as in~\cite{klokov2017escape}. 
During testing, we perform ten voting tests while randomly sampling 1024 points and average the predictions.

\textbf{Point clouds classification.} We report the results of our model and compare it with other approaches in the Table \ref{table:classification}. 
Three different training and testing setting are performed, which are introduced in Section \ref{section:implementation}. 
All approaches, except for the last five which are specially designed for rotation learning, perform well in the Y/Y setting, but experience a significant drop in accuracy when evaluated on unseen rotation as shown in the Y/AR setting. 
We conclude that these approaches only generalize well to rotations that they are trained on. 
However, our proposed method performs equally well across all three settings. 
Every evaluated approach has lower accuracy in the AR/AR setting than the Y/Y setting, except for RI-CNN, SRINet, ClusterNet, and our proposed method. 
This is most likely due to the difficulty of mapping identical objects in different poses to a similar feature space.
This requires larger model complexity and is difficult to address with just dataset augmentation. 
RI-CNN (86.4\%), SRINet (87.0\%) and ClusterNet (87.1\%) are specially designed for rotation invariance. 
Our model has superior performance in the setting of Y/AR and AR/AR (91.0\%), which means our proposed method generalizes well to unseen rotations.

\begin{table}[t]
\begin{center}
\begin{tabular}{lcccc}
\hline
Method & Input size & Y/Y & Y/AR & AR/AR \\
\hline
PointNet~\cite{qi2017pointnet} & $2048 \times 3$ & 79.3 & 43.0 & 73.9 \\
PointNet++~\cite{qi2017pointnet++} & $2048 \times 3$ & \textbf{80.6} & 45.9 & 75.5 \\
DG-CNN~\cite{wang2019dynamic} & $2048 \times 3$ & 79.2 & 46.1 & 71.8 \\
RS-CNN~\cite{liu2019relation} & $2048 \times 3$ & 80.0 & 50.7 & 73.3\\
\hline
RI-CNN~\cite{zhang2019rotation} & $2048 \times 3$ & - & 75.3 & 75.5\\
SRINet~\cite{sun2019srinet} & $2048 \times 3$ & 77.0 & 77.0 & 77.0\\
\hline
Ours & $2048 \times 3$ & 80.2 & \textbf{80.2} & \textbf{80.2} \\
\hline
\end{tabular}
\end{center}
\caption{Part segmentation results on ShapeNet dataset.
Point cooridnates are taken as inputs, and \ac{mIoU} across all classes is reported in three different settings including Y/Y, Y/AR and AR/AR. 
Our model outperforms all approaches except PointNet++ in Y/Y setting. 
Our model has superior performance in the Y/AR and AR/AR settings, which means that it can generalize well to unseen rotations.}
\label{table:segmentation}
\vspace{-2mm}
\end{table}

\begin{figure}[t]
    \centering
    \includegraphics[width=\linewidth,height=0.6\linewidth]{{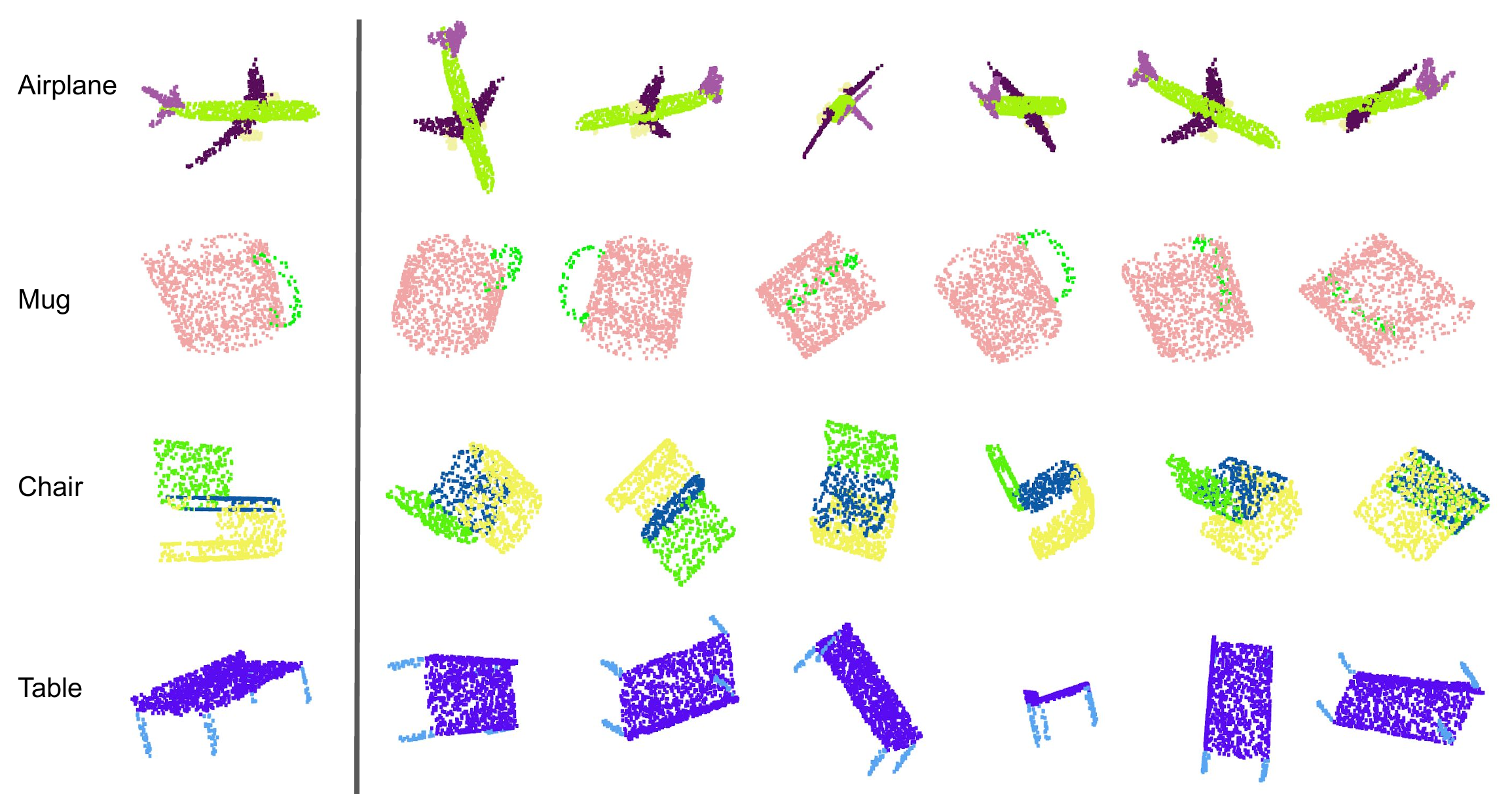}}
    \caption{Qualitative results of our proposed method on part segmentation task on the ShapeNet dataset. 
    From top to bottom, segmentation results of different categories are shown. 
    From left to right, we show ground truth label and results when the input point clouds are arbitrarily rotated during testing.
    Different colors correspond to different part categories. 
    Our model is robust to arbitrary rotations of the input point clouds.}
    \label{fig:seg}
\end{figure}

\begin{figure}[t]
    \centering
    \includegraphics[width=\linewidth,height=0.6\linewidth]{{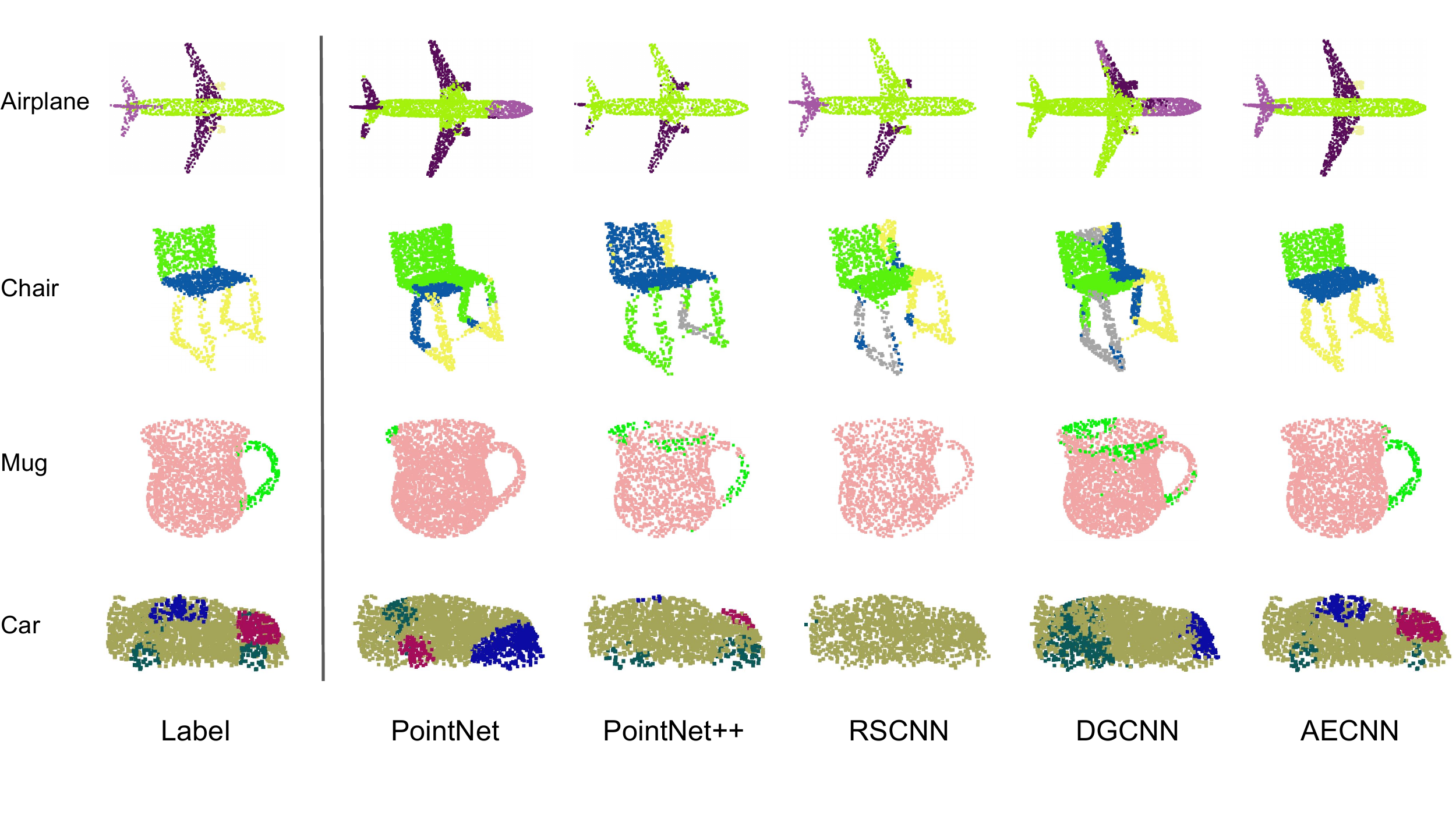}}
    \caption{Qualitative results of part segmentation on ShapeNet. 
    The Y/AR setting is adopted for all models. 
    From top to bottom, segmentation results of different categories are shown. 
    From left to right, we show ground truth label and the results from different approaches. 
    Different colors correspond to different part categories. 
    Our method achieves state-of-the-art performance while other approaches fail to generalize to unseen rotation.}
    \label{fig:seg_unseen_comparison}
\end{figure}

\subsection{Part Segmentation}
\label{section:shapenet}
The part segmentation task requires assigning each point in a point cloud a category label. 
Since this is a point-wise classification task, part segmentation is typically more challenging than classification.

\textbf{Dataset.} We evaluate our model on ShapeNet part dataset~\cite{yi2016scalable}, which contains 16,881 shapes from 16 categories and annotated with 50 parts in total. 
We split the dataset into training, validation and test sets following the convention in PointNet++~\cite{qi2017pointnet++}. 
2048 points are randomly picked on the shape of objects. 
We concatenate the one-hot encoding of the object label to the last feature layer in the model as in~\cite{qi2017pointnet++}. 
During evaluation, mean inter-over-union (mIoU) that are averaged across all classes is reported.

\textbf{3D part segmentation.} We report the result of our model and compare it with other approaches in Table \ref{table:segmentation} and (subfigure a) in the Fig. \ref{fig:seg_comparison}.
Results align well with performance in the classification task. 
In addition, our method outperforms other approaches in all three settings, except for PointNet++ which is slightly better than our proposed method in the Y/Y setting. 
The consistent performance of our method in all three settings demonstrates good generalization to unseen rotations. 
Qualitative results of part segments are illustrated in Fig.~\ref{fig:seg}. 
The comparison results with other approaches in the Y/AR setting are also visualized in Fig.~\ref{fig:seg_unseen_comparison}.

\begin{table}[t]
\small
\begin{center}
\begin{tabular}{lcccc}
\hline
Method & EdgeConv & AEConv1 & AEConv2 & AEConv3\\
\hline
Acc. & 89.6 & 90.2 & 48.5 & 91.0 \\
Para. & 1.94M & 2.14M & - & 1.99M \\
FLOPs & 4170M & 6393M & - & 4841M \\
\hline
\end{tabular}
\end{center}
\caption{Ablation study on aligned edge convolution and comparison with original edge convolution. Accuracy, number of parameters and FLOPs per sample are reported. No further experiments were done on AEConv2 due to its poor accuracy. The number of neighbors is 48. }
\label{table:aligned_edgeconv}
\end{table}

\subsection{\ac{LRF} Analysis}
\label{section:LRF_validation}

We achieve rotation invariance by expressing points coordinates respect to \acp{LRF}. 
Note that the \ac{LRF} is handcrafted rather than learned from raw data. 
As a result, one may be concerned about the effectiveness of the designed \ac{LRF}~\cite{spezialetti2019learning}. 
Currently, common ways of designing \acp{LRF} use the eigenvectors of the covariance matrix of the local point set~\cite{mian2010repeatability,tombari2010unique} or rely on the surface normal as the reference axis~\cite{petrelli2011repeatability}. 
However, computing eignvectors for all points in some local neighborhood of points is time-consuming and estimating an accurate normal from point clouds is still challenging. 
In this study, we consider several alternative ways to compute \acp{LRF} and illustrate their effects on our models' performance. 

The definition of \ac{LRF} is introduced in the Section \ref{section:RIR}. 
Note that given a reference point, the only variation in the definition of the \ac{LRF} arises from the $x$ axis. 
Because the $z$ axis is defined as the direction from the global origin to the reference point, and the $y$ axis is defined by the cross product of the $z$ and $x$ axis. 
Recall that the $x$ axis is associated with the anchor point $m$, which is shown in the Fig. \ref{fig:LRF}. 
Here we investigate three different aspects that influence the determination of the anchor point: searching methods, ways of grouping the data, and the number of neighbors. 
We perform experiments on shape classification, and Table \ref{table:LRF} shows the results.

\begin{table}[t]
\small
\begin{center}
\begin{tabular}{cc|cc|cccc|c}
\hline
\multicolumn{2}{c|}{Searching} & \multicolumn{2}{c|}{Grouping} & \multicolumn{4}{c|}{\# neighbors} & \multirow{2}{*}{Acc.} \\\cline{1-8}
knn & ball & Mean & Max. D & 10 & 16 & 32 & 48 \\
\hline
 & \checkmark & \checkmark &  &  &  &  & \checkmark & 90.4 \\
 & \checkmark &  & \checkmark &  &  &  & \checkmark & 90.3 \\
\checkmark &  &  & \checkmark &  &  &  & \checkmark & 90.3 \\
\checkmark &  & \checkmark &  &  &  &  & \checkmark & \textbf{91.0} \\
\checkmark &  & \checkmark &  & \checkmark &  &  &  & 89.6 \\
\checkmark &  & \checkmark &  &  & \checkmark &  &  & 90.3 \\
\checkmark &  & \checkmark &  &  &  & \checkmark &  & 90.8 \\
\hline
\end{tabular}
\end{center}
\caption{Ablation study on \acp{LRF}. Models are evaluated on ModelNet40 dataset in the  Y/AR setting. Three aspects which effect \acp{LRF} are studied: searching methods, grouping ways and the number of neighbors. }
\label{table:LRF}
\end{table}

We compare two searching methods: ball query and k-\ac{NN} search. 
Ball query finds all points within a certain radius to the reference point, but only up to $k$ points are considered in the experiments. k-\ac{NN} finds a fixed $k$ nearest points to the reference point. 
As is shown in the first four rows of Table \ref{table:LRF}, given the same grouping method models with k-\ac{NN} search have equal or higher accuracy than ball query. 
However the performance of the ball query method is marginally sensitive to the size of the radius chosen, which needs to be assigned manually, as is shown in Table \ref{table:radius}. 

\begin{table}[t]
\begin{center}
\begin{tabular}{ccccc}
\hline
Radii & (0.1, 0.2) & (0.1, 0.4) & (0.2, 0.2) & (0.2, 0.4)\\
\hline
Acc. & 89.1 & 89.3 & 90.3 & \textbf{90.4} \\
\hline
\end{tabular}
\end{center}
\caption{Ablation study on radius in ball query. Models are evaluated on ModelNet40 dataset in the Y/AR setting. Model's performance is marginally sensitive to radius if ball query is used. (r1, r2) indicates the radius in the first SABlock and the second SABlock respectively.}
\label{table:radius}
\end{table}

We study two  ways of determining the anchor point: we define the anchor point as the mean of neighboring points or as the point with largest projected distance to reference point on plane $\alpha$ shown in the Fig. \ref{fig:LRF} (b). 
From rows three and four in Table \ref{table:LRF}, we conclude that anchor points computed from mean of neighbors is preferred (91.0\%) over anchor points with largest projection distance (90.3\%). 
The last four rows of Table \ref{table:LRF} illustrate the performance of our model with different numbers of neighboring points.
We find performance drops with decreasing of $k$ and the model with 48 nearest points achieves the best performance (91.0\%). Further increasing the number of nearest points leads to extra computation burden.

\subsection{Aligned EdgeConv Analysis}
\label{section:aligned_edgeconv}

The proposed AECNN is specially designed to learn a rotation-invariant representation. 
Recall that we align features before doing feature fusion within the edge convolution and call it aligned edge convolution (AEConv).
This study compares the original edge convolution which does not perform feature alignment~\cite{wang2019dynamic}  with our proposed aligned edge convolution. 
We also experiment with different strategies for doing alignment. 
The results are shown in Table \ref{table:aligned_edgeconv}.

We report three different strategies for alignment: transforming the source feature by a transformation matrix in the feature space (AEConv1), which is defined in \eqref{eq:3}; taking source feature $x_j$, LRF $e_i$ of source point, LRF $e_j$ of reference point and translation $T$ as inputs to predict the aligned feature (AEConv2); taking source $x_j$ along with rotation matrix $R$ and translation $T$ as inputs to predict the aligned feature (AEConv3), which is defined in \eqref{eq:4}. 
Our proposed feature alignment idea is verified by comparison between the first column and the last columns, where aligned edge convolution has higher accuracy (91.0 \%) than the original edge convolution (89.6 \%) which has no alignment process.
AEConv1 (90.2 \%) also outperforms edge convolution, but it loses slightly to AEConv3. 
Due to limited GPU memory, we need to reduce the number of learning kernels within the SAFirst block of AEConv1, so that it can be fed into a single GPU during training. 
Even in this case, AEConv1 still has more parameters (2.14M) to learn than AEConv3 (1.99M) and more FLOPs per sample during the test (6393M) than AEConv3 (4841M).
Additionally, AEConv2 is not able to converge.

\section{Conclusions}
This work proposes AECNN, or the Aligned Edge Convolutional Neural Network, which addresses the challenges of learning rotationally-invariant representations for point clouds. 
Rotation invariance is achieved by representing points' coordinates relative to local reference frame. 
The proposed AECNN architecture is designed to better extract and fuse information from local and global features. 
In this way, the AECNN architecture is able to generalize well to unseen rotation. 
Extensive experiments are performed in classification and segmentation and demonstrate effectiveness of AECNN. 

\section{Acknowledgement}
This work was supported by a grant from the Ford Motor Company via the Ford–University of Michigan Alliance under award N028603.

{\small
\bibliographystyle{ieee}
\bibliography{egbib}
}

\end{document}